\documentclass[journal,twoside,web]{ieeecolor}
\usepackage{tmi}
\usepackage{cite}
\usepackage{amsmath,amssymb,amsfonts}
\usepackage{algorithmic}
\usepackage{graphicx}
\usepackage{textcomp}
\usepackage{booktabs}
\usepackage{colortbl}
\usepackage{multirow}
\usepackage{url}
\usepackage{float}
\definecolor{mygray}{RGB}{230, 230, 230}
\definecolor{myorigin}{RGB}{255, 231, 154}
\definecolor{myblue}{RGB}{189, 215, 238}
\definecolor{dino}{RGB}{249,231,227}
\def\BibTeX{{\rm B\kern-.05em{\sc i\kern-.025em b}\kern-.08em
    T\kern-.1667em\lower.7ex\hbox{E}\kern-.125emX}}
\begin{document}
\title{SurgPETL: Parameter-Efficient Image-to-Surgical-Video Transfer Learning for Surgical Phase Recognition}
\author{Shu Yang, Zhiyuan Cai, Luyang Luo, Ning Ma, Shuchang Xu and Hao Chen, \IEEEmembership{Senior Member, IEEE}
\thanks{ This work was supported by in part by the Hong Kong Innovation and Technology Fund (Project No. GHP/006/22GD), the Project of Hetao Shenzhen-Hong Kong Science and Technology Innovation Cooperation Zone (HZQB-KCZYB-2020083), the Research Grants Council of the Hong Kong (Project Reference Number: T45-401/22-N), NSFC General Project (62072452), and the Regional Joint Fund of Guangdong under Grant (2021B1515120011). \emph{(Corresponding author: Hao Chen.)}}
\thanks{Shu Yang, Zhiyuan Cai and Luyang Luo, Ning Ma are with the Department of Computer Science and Engineering, Hong Kong University of Science and Technology, Hong Kong, China (e-mail: syangcw@connect.ust.hk, zcaiap@connect.ust.hk, cseluyang@ust.hk, csemaning@ust.hk).}
\thanks{Shuchang Xu is with the Department of Gastroenterology, Tongji Institute of Digestive Disease, Tongji Hospital, School of Medicine, Tongji University, Shanghai, China (e-mail: xschang@163.com).}
\thanks{Hao Chen is with the Department of Computer Science and Engineering, the Department of Chemical and Biological Engineering, the Division of Life Science, Hong Kong University of Science and Technology, Hong Kong, China and HKUST Shenzhen-Hong Kong Collaborative Innovation Research Institute, Futian, Shenzhen, China (e-mail: jhc@cse.ust.hk).}
}

\maketitle

\begin{abstract}
Capitalizing on image-level pre-trained models for various downstream tasks has recently emerged with promising performance. However, the paradigm of ``image pre-training followed by video fine-tuning” for high-dimensional video data inevitably poses significant performance bottlenecks. Furthermore, in the medical domain, many surgical video tasks encounter additional challenges posed by the limited availability of video data and the necessity for comprehensive spatial-temporal modeling. Recently, Parameter-Efficient Image-to-Video Transfer Learning (PEIVTL) has emerged as an efficient and effective paradigm for video action recognition tasks, which employs image-level pre-trained models with promising feature transferability and involves cross-modality temporal modeling with minimal fine-tuning. Nevertheless, the effectiveness and generalizability of this paradigm within intricate surgical domain remain unexplored. In this paper, we delve into a novel problem of efficiently adapting image-level pre-trained models to specialize in fine-grained surgical phase recognition, termed as Parameter-Efficient Image-to-Surgical-Video Transfer Learning. Firstly, we develop a parameter-efficient transfer learning benchmark SurgPETL for surgical phase recognition, and conduct extensive experiments with three advanced methods based on ViTs of two distinct scales pre-trained on five large-scale natural and medical datasets. Then, we introduce the Spatial-Temporal Adaptation (STA) module, integrating a standard spatial adapter with a novel temporal adapter to capture detailed spatial features and establish connections across temporal sequences for robust spatial-temporal modeling. Extensive experiments on three challenging datasets spanning various surgical procedures demonstrate the effectiveness of SurgPETL with STA. SurgPETL-STA outperforms both parameter-efficient alternatives and state-of-the-art surgical phase recognition methods while maintaining parameter efficiency and minimizing overhead. 
\end{abstract}


\section{Introduction}
\label{sec:introduction}
\IEEEPARstart{T}{he} pre-training and fine-tuning paradigm has emerged as a dominant technique in the field of transfer learning. The \textit{de facto} standard approaches typically adopt image-level pre-trained models for initialization, followed by fine-tuning all parameters, yielding impressive performance. Nevertheless, with the advent of massively pre-trained foundation models, this routine practice becomes increasingly prohibitive and computationally infeasible. Moreover, processing the high-dimensional video data further exacerbates the computational cost of such paradigm. In contrast, parameter-efficient transfer learning (PETL)~\cite{Clip-adapter,Prefix-tuning} has demonstrated favorable performance with remarkable computational efficacy by fine-tuning a few parameters while keeping pre-trained models frozen. Inspired by PETL, recent methods~\cite{ST-Adapter,AIM,DUAL-PATH} tailored for video action recognition tasks exploit large pre-trained image models with sufficient knowledge and selectively fine-tune a few parameters for spatial-temporal reasoning, namely parameter-efficient image-to-video transfer learning (PEIVTL).

\begin{figure*}[t]
    \centering
    \begin{minipage}[t]{0.65\linewidth}
        \centering
        \includegraphics[width=\linewidth]{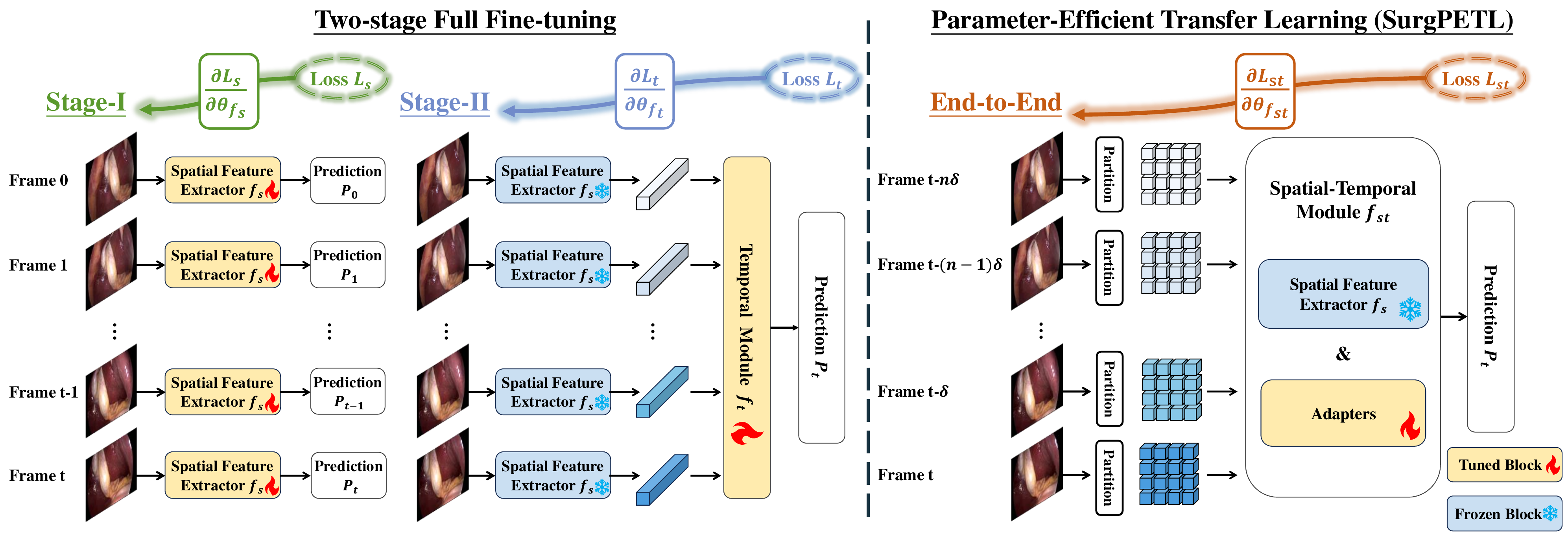}
        \caption{Illustration of two distinct paradigms for surgical phase recognition. \textbf{Left}: two-stage full fine-tuning paradigm; \textbf{Right}: end-to-end parameter-efficient image-to-surgical-video transfer learning paradigm.}
        \label{fig:motivation}
    \end{minipage}
    \hfill
    \begin{minipage}[t]{0.34\linewidth}
        \centering
        \includegraphics[width=\linewidth]{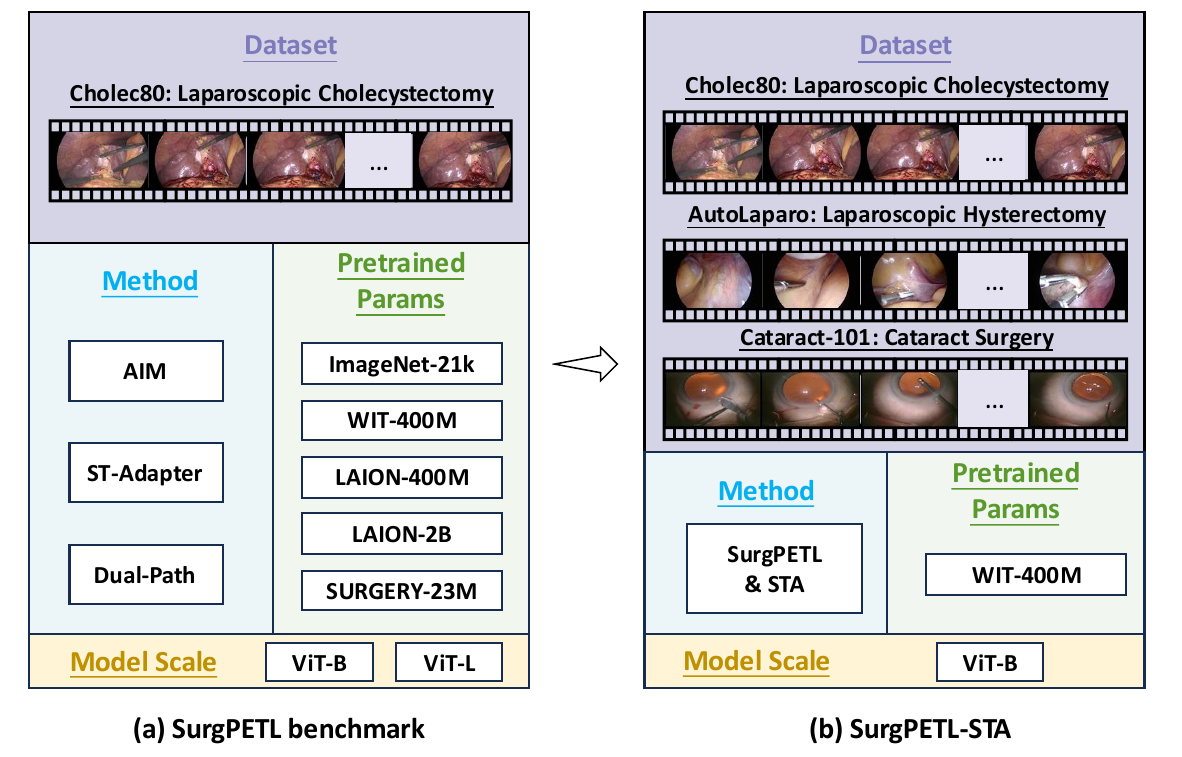}
        \caption{Illustration of the investigation with two distinct phases for parameter-efficient image-to-surgical-video transfer learning.}
        \label{fig:pipeline}
    \end{minipage}
\end{figure*}

Despite the impressive performance in video action recognition tasks, PEIVTL remains underexplored in surgical phase recognition tasks, which is primarily attributed to the inherent discrepancies in learning objectives and specialized domain knowledge. \textbf{Video action recognition} aims to automatically assign action categories to individual \textit{natural videos}, with an emphasis on \textit{video-level analysis}. 
In contrast, \textbf{surgical phase recognition} focuses on automatically assigning phase labels to each frame of \textit{surgical footage}, emphasizing \textit{fine-grained frame-level analysis}. Due to the temporal ambiguities between distinct surgical phases, precise phase recognition necessitates a comprehensive understanding of scenes over time, along with the adequate integration of both spatial details and temporal dynamics. However, current methods focus on capturing coarse global temporal information~\cite{AIM,ST-Adapter,DUAL-PATH}, failing to provide more advanced temporal modeling capabilities for fine-grained tasks. Consequently, extending efficient transfer learning to the surgical domain becomes more challenging. 

In this work, we investigate a novel and critical problem of efficiently adapting pre-trained image models to specialize in fine-grained surgical phase recognition, namely parameter-efficient image-to-surgical-video transfer learning. In addition to the fundamental motivation inherent in PEIVTL, this investigation is driven by two extra special factors:

\textbf{1)} Datasets specifically designed for the surgical domain are often disproportionately small due to the heavy reliance on manual annotations from clinical experts. Efficiently adapting pre-trained image models can provide a solid foundation for fine-tuning with limited annotated datasets to mitigate the overfitting issues posed by fully fine-tuning on restricted surgical datasets, therefore achieving enhanced performance.

\textbf{2)} The two-stage paradigm has emerged as the prevailing method for surgical phase recognition~\cite{LoViT,SKiT}, sequentially extracting spatial and temporal information as depicted on the \textit{left} of Fig.~\ref{fig:motivation}. Besides its high training cost, this paradigm impedes joint spatial-temporal modeling and underutilizes the potential of the pre-existing image-based knowledge.

Motivated by the above observations, our investigation is organized into two distinct phases, as illustrated in Fig.~\ref{fig:pipeline}, to transfer the success of PEIVTL from the natural domain to the surgical domain. \textbf{Firstly}, as shown on the \textit{right} of Fig.~\ref{fig:motivation}, we propose an end-to-end framework, termed as SurgPETL, which employs a frozen spatial feature extractor alongside learnable adapters for efficient spatial-temporal modeling. Specifically, we implement SurgPETL for surgical phase recognition by conducting comprehensively experiments with three state-of-the-art methods, based on ViTs of two distinct scales pre-trained on five different large-scale datasets (as depicted in Fig.~\ref{fig:pipeline} (a)). 
\textbf{Then}, we introduce the Spatial-Temporal Adaptation (STA) module to integrate two distinct spatial and temporal adapters, tailored specially for effective spatial-temporal modeling across both spatial and temporal dimensions. Distinct from common adapters tailored for spatial information, the proposed temporal adapter promotes frame collaboration and facilitates the learning of implicit motion information among frames, thereby improving spatial-temporal modeling. Meanwhile, we employ a spatial-temporal router to balance the contributions of re-embedded spatial and temporal features from the two adapters, which suppresses the redundant and misleading information to capture more discriminative information. To evaluate the effectiveness of the proposed SurgPETL and STA, we conduct comprehensive experiments on three public challenging datasets spanning distinct surgical procedures, as shown in Fig.~\ref{fig:pipeline} (b).

Our contributions can be summarized as follows. (1) We delve into a novel and significant problem, termed as parameter-efficient image-to-surgical-video transfer learning, and establish the SurgPETL benchmark for surgical phase recognition with extensive experiments. (2) Our proposed SurgPETL simplifies the two-stage pipeline of surgical phase recognition and effectively tackles the challenges associated with inadequate spatial-temporal modeling and overfitting. (3) We propose the Spatial-Temporal Adaption (STA) module with one foundational spatial adapter and one novel temporal adapter, tailored specially for effective spatial-temporal modeling. (4) We conduct extensive experiments on three challenging datasets and verify that our proposed SurgPETL with STA outperforms existing parameter-efficient alternatives and state-of-the-art surgical phase recognition models.

\section{Related Work}

\subsection{Image Pre-trained Models}
ViT~\cite{ViT} and its variants~\cite{swintransformer,PVT,T2T-VIT} have demonstrated impressive generalizability and scalability in their feature representations. These pre-trained image models offer rich and reliable spatial feature representations, making them widely adopted as effective initializations for various computer vision tasks, such as segmentation~\cite{zhang2016saliency,zhang2018bi,yang2023real} and detection~\cite{huang2018yolo}. The emergence of multi-modality techniques~\cite{ALIGN,CLIP,BEIT-3} and the availability of large-scale web-collected image-text pair datasets (e.g., LAION-5B~\cite{Laion-5B}) have further facilitated the learning of powerful and discriminative visual representations.

In the surgical domain, conventional fully-supervised learning methods necessitate extensive annotated data, imposing prohibitively high costs. Consequently, several methods~\cite{endovit,free,Surgery-23M,SelfSupSurg} focus on exploring the effectiveness of self-supervised learning (SSL) in the complex surgical domain. Early methods~\cite{free,SelfSupSurg} employ SSL on small-scale public datasets to enhance the understanding of surgical contexts. SelfSupSurg~\cite{SelfSupSurg} presents a comprehensive analysis of four advanced SSL methods (MoCo v2~\cite{MoCo_v2}, SimCLR~\cite{SimCLR}, DINO~\cite{DINO}, SwAV~\cite{SwAV}) across a range of surgical tasks. Several recent approaches~\cite{endovit,Surgery-23M} develop image-level endoscopy-specific pre-trained models using lager-scale datasets. For instance, EndoViT~\cite{endovit} assembles the largest publicly available corpus of endoscopic images and develops a foundation model specifically pre-trained on this dataset by employing advanced MAE~\cite{MAE} method. EndoSSL~\cite{Surgery-23M} constructs extensive unlabeled endoscopic video datasets to train an image-level foundation model using MSN~\cite{MSN}, resulting in enhanced representational capabilities. In this work, we leverage mainstream open-source image-level parameters pre-trained on natural or surgical domains and adapt them for surgical phase recognition with minimal fine-tuning.

\subsection{Parameter-Efficient Transfer Learning}
Efficient transfer learning has demonstrated its effectiveness and efficiency in both natural language processing and image-level computer vision tasks. Parameter-efficient transfer learning paradigms, such as Adapters~\cite{Adapters}, Prefix Tuning~\cite{Prefix-tuning} and LoRA~\cite{Lora}, focus on fine-tuning additional parameters while keeping pre-trained models frozen. However, the rapid growth of video applications in diverse fields~\cite{yang2021learning,shafiee2017fast,zhang2019fast} results in increased computational demands. To address these challenges, parameter-efficient image-to-video transfer learning involves the complex task of bridging the gap between spatial and temporal dimensions, particularly the challenge of encoding the temporal context of videos by leveraging the discriminative spatial information provided by pre-trained image models. EVL~\cite{EVL} employs a lightweight transformer decoder to aggregate frame-level spatial features from the frozen encoder. ST-Adapter~\cite{ST-Adapter} sequentially integrates two depth-wise 3D convolutional layers to enable spatial-temporal reasoning. AIM~\cite{AIM} introduces spatial, temporal and joint adaptations through three adapters with identical architectures, progressively equipping an image model with spatial-temporal reasoning capabilities. Dual-Path~\cite{DUAL-PATH} employs a pre-trained image model to process a grid-like frameset for capturing temporal information, while simultaneously analyzing individual frames to extract spatial information. Despite their impressive performance, these methods primarily focus on coarse global temporal information for video-level analysis, and struggle to generate discriminative temporal information necessary for fine-grained surgical phase recognition. In this work, we address this limitation by introducing a novel Spatial-Temporal Adaptation module to efficiently adapt pre-trained models for specialized fine-grained surgical phase recognition.

\subsection{Surgical Phase Recognition}
Automatic surgical phase recognition aims to assign phase labels to individual frames within surgical videos~\cite{TeCNO, Trans-SVNet, LoViT, SKiT, guo2024surgical, yang2024surgformer}. The field has been dominated by the two-stage learning paradigm, which involves the sequential extraction of spatial and temporal features. Numerous research efforts leverage pre-trained image models for initializing spatial feature extractor and introduce advanced temporal modules to capture temporal features. For instance, TeCNO~\cite{TeCNO} utilizes a multi-stage temporal convolutional network~\cite{MSTCN} for hierarchical prediction refinement, while Trans-SVNet~\cite{Trans-SVNet} integrates spatial and temporal features as hybrid embedding and enhance their synergy via transformer. SkiT~\cite{SKiT} captures critical information through an efficient key pooling operation with bounded time complexity, which maintains consistent inference times across varied lengths. Despite these advances, the two-stage pipeline impedes the joint spatial-temporal modeling and effective utilization of pre-trained image models. In this work, we propose an end-to-end framework that effectively utilizes pre-trained image models with minimal parameter tuning, significantly reducing computational overhead while addressing challenges in joint spatial-temporal modeling and mitigating overfitting.

\section{Methodology}
In this section, we start by presenting the preliminaries related to the end-to-end paradigm for surgical phase recognition in~$\S$\ref{Sec:1}. Subsequently, we elaborate on two distinct phases of our investigation in~$\S$\ref{Sec:2} and the selected method benchmarks in~$\S$\ref{Sec:3}. Furthermore, in~$\S$\ref{Sec:4}, we introduce our proposed SurgPETL-STA, which integrates the SurgPETL framework with Spatial-Temporal Adaption (STA) module.

\subsection{Preliminaries}
\label{Sec:1}
Initially, we present a recapitulation of the transformer architecture, as exemplified by ViT~\cite{ViT}. ViT stacks several blocks to extract feature representations, where each block comprises two key components: multi-head self-attention (MSA) and multi-layer perceptron (MLP). Given an input image $I\in\mathbb{R}^{C\times H\times W}$, ViT first partitions the image into non-overlapping patches of size $P \times P$. These patches are then mapped to embedding $X\in\mathbb{R}^{K\times D}$, where $K=HW/P^2$ and $D$ denotes the latent vector size. The block can be formulated as follows:
\begin{equation}
\begin{aligned}
    X'& = X + MSA(LN(X)), \\
    X''& = X' + MLP(LN(X')),
\end{aligned}
\end{equation}
where $LN(\cdot)$ refers to the layer normalization operation, $MSA(\cdot)$ indicates the multi-head self-attention operation, and $MLP(\cdot)$ contains two linear layers with a GELU.

Subsequently, we briefly describe how to apply transformer architecture to surgical phase recognition in an end-to-end manner. Given untrimmed surgical footage, a sparsification strategy is employed to generate a frame volume $V\in\mathbb{R}^{T\times C\times H\times W}$ for each frame, with the objective of predicting the surgical phase of corresponding frame. Specifically, we sample $T$ frames from the surgical video at a frame rate $R$, which is driven by the observation that the motions within local neighboring frames exhibit subtle variations. By carefully selecting values for $T$ and $R$, we can effectively capture essential temporal information, tackling ambiguous transitions while minimizing spatial-temporal redundancy across adjacent frames. For each frame, the related frame volume $V$ is divided into non-overlapping patches of size $1 \times P\times P$, arranged in spatial-temporal order. These patches are then mapped to spatial-temporal tokens $X\in\mathbb{R}^{T\times K\times D}$, which are subsequently fed into transformer architecture for phase prediction. For PEIVTL, mainstream methods utilize MSA primarily in the spatial domain to prevent quadratic growth in computational cost as the sequence length $T$ expands. However, effective temporal modeling is critical for accurate surgical phase recognition, as it requires robust spatial-temporal features to handle the complex dynamics inherent in surgical procedures. Therefore, a significant challenge lies in effectively extracting the temporal context from videos by leveraging the discriminative spatial context provided by pre-trained image models.

\subsection{Two-phase Investigation}
\label{Sec:2}
In this section, we present the principal pipeline, structured into two pivotal phases aimed at investigating the effectiveness of parameter-efficient image-to-surgical-video transfer learning. As illustrated in Fig.~\ref{fig:pipeline}, the two distinct phases encompass: 1) SurgPETL benchmark; 2) SurgPETL-STA.

In \textbf{Phase-I}, our investigation systematically examines the adaptation of pre-trained image models for surgical phase recognition by utilizing parameter-efficient image-to-video transfer learning paradigm. To achieve this, we establish the SurgPETL benchmark incorporating three advanced methods, based on ViTs of two different scales pre-trained on five diverse large-scale datasets.

\noindent\textbf{Methods.} To ensure a thorough assessment of the latest advancements, we select three PEIVTL methods tailored for natural video-level tasks, including AIM~\cite{AIM}, ST-Adapter~\cite{ST-Adapter} and Dual-Path~\cite{DUAL-PATH}, as elaborated in $\S$\ref{Sec:benchmark}.

\noindent\textbf{Scales.} Our objective is to access the overall scalability of these methods across different model capacities, thereby verifying their generalizability in the surgical domain. Specifically, we utilize two model scales: ViT-B, which comprises 12 transformer blocks with 86 million parameters, and ViT-L, a larger model with 24 blocks and 307 million parameters.

\noindent\textbf{Pre-trained Parameters.} To assess the general transferability of these methods, we employ the parameters pre-trained on distinct datasets from the natural and surgical domains:
\begin{itemize}
\item ImageNet-21K~\cite{imagenet} dataset contains 1.4 billion images annotated with 21,000 classes. The parameters are derived from a full-supervised pre-training process on this dataset.

\item WIT-400M~\cite{CLIP} dataset comprises a collection of 400 million image-text pairs collected from the internet. The parameters are obtained through a self-supervised training procedure conducted on this dataset, adapting CLIP.

\item LAION-400M~\cite{Laion-400M} dataset comprises 400 million image-text pairs filtered via CLIP preprocessing. The parameters are derived from a self-supervised pre-training process utilizing CLIP on this dataset.

\item LAION-2B~\cite{Laion-5B} contains 2.32 billion English image-text pairs. The parameters are derived from a self-supervised training process on this dataset with CLIP.

\item SURGERY-23M~\cite{Surgery-23M} dataset consists of 7,877 laparoscopic procedures videos, encompassing 23.3 million frames. The parameters are acquired from a pre-training process utilizing MSN~\cite{MSN} on this private dataset. Note that the parameters based on ViT-B are not available.
\end{itemize}

In \textbf{Phase-II}, we introduce the Spatial-Temporal Adaption (STA) module, which integrates a spatial adapter with a novel temporal adapter to capture detailed spatial features and establish connections across temporal sequences for effective temporal modeling. Building on the SurgPETL benchmark, STA module is seamlessly incorporated into the outputs of temporal attention within each block to boost the spatial-temporal modeling capabilities of the pre-trained image models (refer to $\S$\ref{Sec:4} for further details). As illustrated in Fig.~\ref{fig:pipeline} (b), we perform extensive experiments on three public challenging datasets spanning various surgical procedures. For efficient fine-tuning and fair comparisons with existing methods, we utilize the ViT-B pre-trained on WIT-400M for initiation.

\subsection{Selected Method Benchmarks}
\label{Sec:3}

\begin{figure}[t]
\centering
\includegraphics[width=0.95\linewidth]{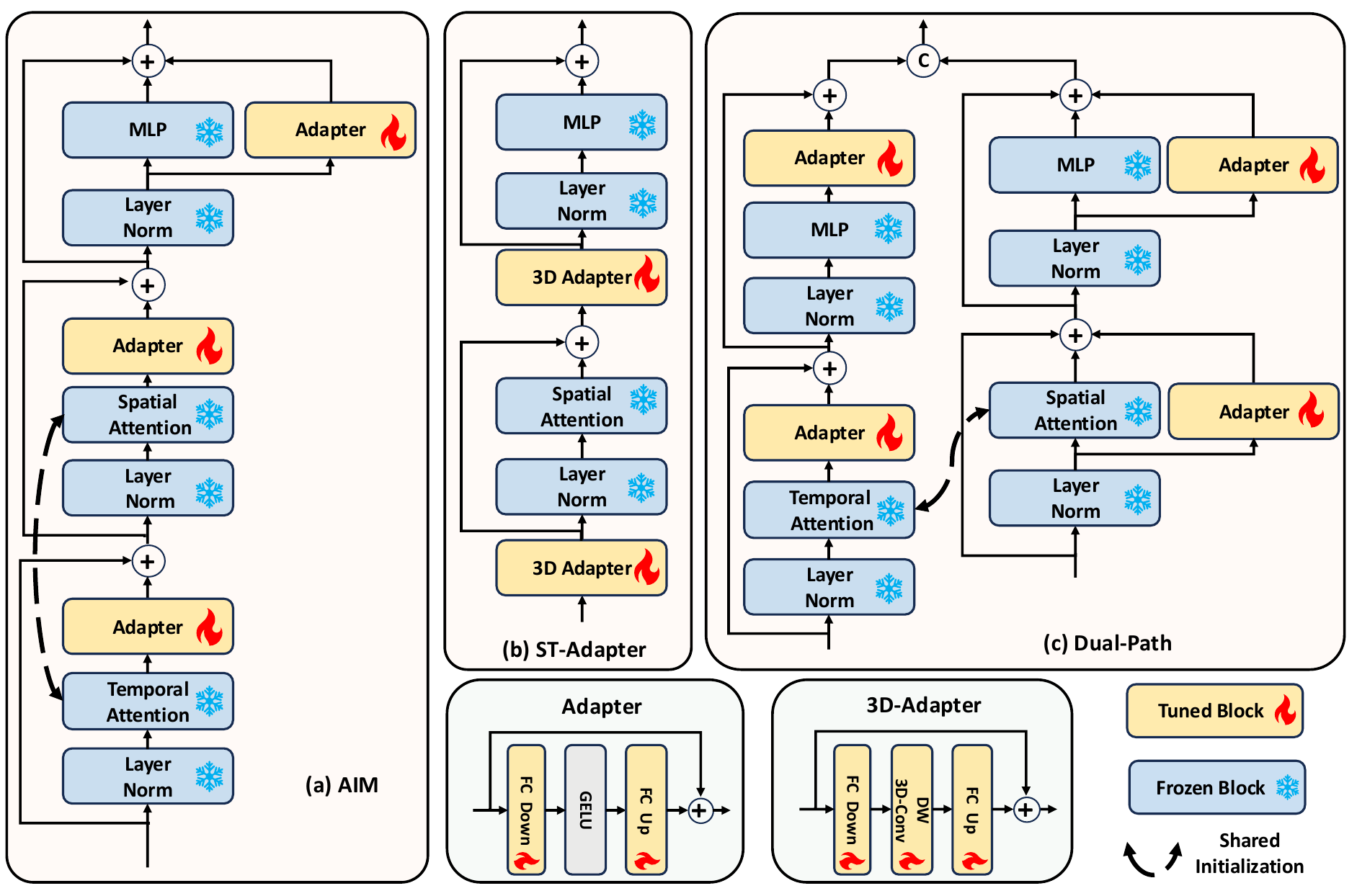}
\caption{Illustration of the selected PEIVTL method benchmarks.}
\label{fig:benchmark}
\end{figure}

\begin{figure*}[t]
\centering
\includegraphics[width=0.75\linewidth]{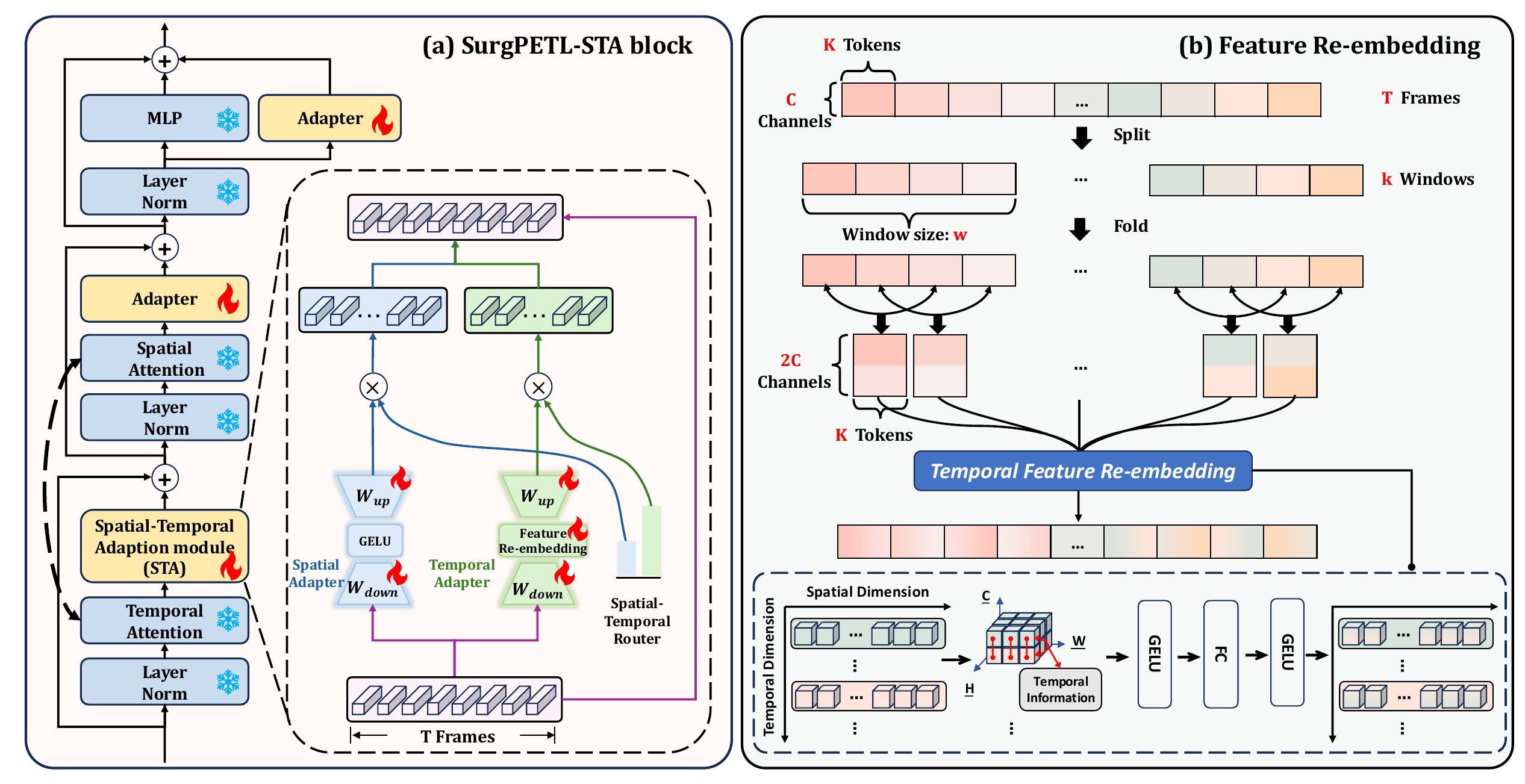}
\caption{Overview of SurgPETL-STA. \textbf{(a)}: SurgPETL-STA block consists of one Spatial-Temporal Adaption (STA) module and two adapters for temporal, spatial and joint adaptations. Each STA contains a spatial adapter, a novel temporal adapter and a spatial-temporal router. \textbf{(b)}: Feature Re-embedding (FR) mechanism of the temporal adapter is utilized to fully utilize the potential temporal feature space for enhanced representation.}
\label{fig:adapter}
\end{figure*}

\label{Sec:benchmark}
\setlength{\fboxsep}{1pt}
Recent advances~\cite{AIM,ST-Adapter,DUAL-PATH} in PEIVTL predominantly leverage the adapter-based techniques, which seamlessly integrate compact adapters within transformer blocks to facilitate the extraction of spatial and temporal feature representations. We systematically analyze notable distinctions among diverse methods and comprehensively describe the representation of spatial and temporal features within these methods. Furthermore, we re-frame these methods as modifications to several specific variables and functions within the formulated equations, thus establishing a coherent framework for understanding and analysis. We represent \colorbox{myorigin}{learnable} and \colorbox{myblue}{frozen} parameters in yellow and blue colors, respectively. We adopt a unified perspective to abstract and conceptualize the adaption mechanism for both the spatial and temporal domains:
\begin{equation}
Out = In' + \colorbox{myorigin}{$\phi$}(In), ~\colorbox{myorigin}{$\phi$}(In) = In + f(In\colorbox{myorigin}{$W_{d}$})\colorbox{myorigin}{$W_{u}$},
\label{eq:spatial}
\end{equation}
where $Out$ is the target output, $In$ and $In'$ denote the input associated with spatial and temporal features. \colorbox{myorigin}{$\phi$} denotes the specific adapter, which is a bottleneck architecture~\cite{Adapters} consisting of two fully connected (FC) layers with a method-specific function $f(\cdot)$ in the middle. \colorbox{myorigin}{$W_{d}$} and \colorbox{myorigin}{$W_{u}$} refer to the parameters of two FC layers for down-projection and up-projection, which project the input to a lower-dimensional space and a higher-dimensional space, respectively.

Given $X\in\mathbb{R}^{T\times K \times D}$, we provide method-specific descriptions for our selected methods, as illustrated in the Fig.~\ref{fig:benchmark}.

\noindent\textbf{AIM} employs a sequential paradigm to derive spatial-temporal feature representations by successively processing temporal and spatial features. It performs temporal, spatial and joint adaptations through three adapters with identical architectures~\cite{Adapters} to gradually equip an image model with spatial-temporal reasoning capability. First, AIM reuses the pre-trained MSA for the temporal dimension, termed T-MSA, and incorporates an adapter into the outputs of temporal attention in each block. This operations could be formulated as,
\begin{equation}
\label{AIM}
X_{t} = X + \colorbox{myorigin}{$\phi$}(\colorbox{myblue}{$T\text{-}MSA$}(\colorbox{myblue}{$LN$}(X))).
\end{equation}
Next, the temporally enhanced feature is fed into the subsequent MSA, tailored for the spatial dimension, with a spatial adapter applied as follows,
\begin{equation}
X_{s} =  X_t + \colorbox{myorigin}{$\phi$}(\colorbox{myblue}{$MSA$}(\colorbox{myblue}{$LN$}(X_t))).
\end{equation}
Finally, the feature is enhanced through the MLP with a joint adaptation via an additional adapter,
\begin{equation}
    X_{st}= X_{s} + \colorbox{myblue}{$MLP$}(\colorbox{myblue}{$LN$}(X_{s})) + \colorbox{myorigin}{$\phi$}(\colorbox{myblue}{$LN$}(X_{s})). \\
\end{equation}
Specifically, the adapter consists of two FC layers and a GELU activation layer in the middle, formulated as:
\begin{equation}
    \colorbox{myorigin}{$\phi$}(In) = \colorbox{myorigin}{$Adapter$}(In) = In + GELU(In\colorbox{myorigin}{$W_{d}$})\colorbox{myorigin}{$W_{u}$}.\\
    \label{adapter}
\end{equation}

\noindent\textbf{ST-Adapter} integrates a depth-wise 3D convolution layer into adapter for spatial-temporal modeling, termed as $3D\text{-}Adapter$. Two $3D\text{-}Adapter$ modules are placed sequentially within each transformer block. The first adapter is positioned before the MSA, while the second is applied after the MSA, allowing for effective spatial-temporal feature extraction throughout the block. The entire function could be formulated as,
\begin{equation}
     X'=\colorbox{myorigin}{$\phi$}(X), X''=\colorbox{myorigin}{$\phi$}(X'+\colorbox{myblue}{$MSA$}(\colorbox{myblue}{$LN$}((X')))).
\end{equation}
Then, the feature is further enhanced through the combination of MLP and LN for refined spatial-temporal modeling,
\begin{equation}
    X_{st} =X'' + \colorbox{myblue}{$MLP$}(\colorbox{myblue}{$LN$}(X'')).\\
\end{equation}
Specifically, $3D\text{-}Adapter$ comprises two FC layers with a depth-wise 3D convolutional layer in between to fill the intrinsic gap between image data and video data,
\begin{equation}
    \colorbox{myorigin}{$\phi$}(In) = \colorbox{myorigin}{$3D\text{-}Adapter$}(In) = In + \colorbox{myorigin}{$Conv3D$}(In\colorbox{myorigin}{$W_{d}$})\colorbox{myorigin}{$W_{u}$}.\\
\end{equation}

\noindent\textbf{Dual-Path} employs a parallel paradigm to aggregate spatial and temporal features from separated spatial and temporal adaptation paths into a unified spatial-temporal representations. For temporal adaptation path, it organizes consecutive frames into a grid-like frameset to generate spatial-temporal tokens $X_t$, which are then fed into blocks with adapters placed on top of the MSA and MLP, enabling the model to learn relationships between local patches across frames:
\begin{equation}
\begin{aligned}
X_{t}' &= X_t + \colorbox{myorigin}{$\phi$}(\colorbox{myblue}{$MSA$}(\colorbox{myblue}{$LN$}(X_t))), \\
X_{t}'' &= X_t' + \colorbox{myorigin}{$\phi$}(\colorbox{myblue}{$MLP$}(\colorbox{myblue}{$LN$}(X_t'))).
\end{aligned}
\end{equation}
For spatial adaption path, it incorporates adapters into the side-outputs of MSA and MLP in each block to utilize the outstanding ability of spatial modeling. The entire function could be formulated as:
\begin{equation}
\begin{aligned}
X_{s}' &=  X_s + \colorbox{myblue}{$MSA$}(\colorbox{myblue}{$LN$}(X_s)) + \colorbox{myorigin}{$\phi$}(\colorbox{myblue}{$LN$}(X_s)), \\
X_{s}'' &=  X_s' + \colorbox{myblue}{$MLP$}(\colorbox{myblue}{$LN$}(X_s')) + \colorbox{myorigin}{$\phi$}(\colorbox{myblue}{$LN$}(X_s')).
\end{aligned}
\end{equation}
Specifically, \colorbox{myorigin}{$\phi$} has the same architecture as that used in AIM. Finally, it devises the aggregation operation as a concatenation of the spatial feature $X_s''$ and temporal feature $X_t''$.
\subsection{SurgPETL-STA}
\label{Sec:4}

In this work, we primarily follow the mainstream framework of AIM, since the experimental results in $\S$\ref{Sec:phase1} demonstrate that AIM is well-suited for surgical phase recognition due to its simplicity and efficiency. Despite the impressive performance, AIM fails to explicitly address the temporal modeling issue. Specifically, the T-MSA component in AIM is implemented by reusing the MSA, which is originally tailored for spatial modeling and remains frozen for minimal fine-tuning. Consequently, the direct employment of T-MSA tailored for spatial dimensions in temporal modeling fails to adequately capture the intricate temporal information in complex scenes. AIM utilizes learnable adapters to learn temporal representations. However, the adapter in \eqref{adapter} tailored for spatial dimensions optimizes each frame independently, thus neglecting the temporal context necessary for effective temporal modeling and diminishing temporal awareness. 

To alleviate this, we propose a novel Spatial-Temporal Adaption (STA) module that conduct feature enhancement across the spatial and temporal dimensions by joint employing a spatial adapter and a novel temporal adapter, capable of modeling robust and discriminative spatial and temporal context. By implementing the STA module for joint spatial-temporal modeling, the variant SurgPETL-STA can learn the temporal relationships among frames, facilitating temporal feature learning. This enables a frozen, reused temporal MSA to effectively reason about complex temporal relations among both adjacent and distant frames in a parameter-efficient manner. The computation of the SurgPETL-STA block can be written as
\begin{equation}
    X_{t}= X + \colorbox{myorigin}{$STA$}(\colorbox{myblue}{$T\text{-}MSA$}(\colorbox{myblue}{$LN$}(X))), \\
\end{equation}
\begin{equation}
    X_{s}= X_t + \colorbox{myorigin}{$Adapter$}(\colorbox{myblue}{$MSA$}(\colorbox{myblue}{$LN$}(X_t))), \\
\end{equation}
\begin{equation}
    X_{st}= X_{s} + \colorbox{myblue}{$MLP$}(\colorbox{myblue}{$LN$}(X_{s})) + \colorbox{myorigin}{$Adapter$}(\colorbox{myblue}{$LN$}(X_{s})). \\
\end{equation}

To specifically address the issues of standard adapters~\cite{Adapters} in spatial-temporal modeling, STA integrates two distinct adapters designed to enhance features along both spatial and temporal dimensions, which compensates for the lack of comprehensive spatial-temporal modeling in standard adapters. Given the temporally enhanced feature $H$, derived from T-MSA and LN, STA first utilizes two specialized adapters: a spatial adapter to refine and enhance the spatial details within each frame and a temporal adapter to strengthen the temporal correlation across frames in the sequence. Independent spatial and temporal adapters complement each other, allowing the model to more effectively capture both intra-frame spatial structure and inter-frame temporal dependencies from original feature representations. Following this, the Spatial-Temporal Router balances the contributions of re-embedded spatial and temporal features to suppress the redundant and misleading information and highlight discriminative features. Finally, STA incorporates a residual structure to combine original input with enhanced spatial-temporal feature, preserving knowledge from the pre-trained image model. This ensures that the essential information from the original input remains intact, while the enhanced features add further refinement of spatial-temporal information. The entire STA function can be formulated as
\begin{equation}
H=\colorbox{myblue}{$T\text{-}MSA$}(\colorbox{myblue}{$LN$}(X)),
\end{equation}
\begin{equation}
    H' = H + \colorbox{myorigin}{$\alpha$}\cdot\underbrace{\colorbox{myorigin}{$g_s$}(H\colorbox{myorigin}{$W_{d}^s$})\colorbox{myorigin}{$W_{u}^s$}}_{\text{Spatial Adapter}} + \colorbox{myorigin}{$\beta$}\cdot\underbrace{\colorbox{myorigin}{$g_t$}(H\colorbox{myorigin}{$W_{d}^t$})\colorbox{myorigin}{$W_{u}^t$}}_{\text{Temporal Adapter}},
\end{equation}
where $\alpha$ and $\beta$ indicate the learnable parameters associated with the Spatial-Temporal Router, governing the aggregation of spatial and temporal features. The spatial adapter utilizes the parameters $W_{d}^s$ and $W_{u}^s$ for its operations, while the temporal adapter operates using $W_{d}^t$ and $W_{u}^t$. Specifically, $W_{d}^s$ and $W_{d}^t$ reduce the dimensionality of the input, and $W_{u}^s$ and $W_{u}^t$ restore it back to the original size. $g_s(\cdot)$ refers to $GELU(\cdot)$ in standard spatial adapter, and $g_t(\cdot)$ denotes the core component in the temporal adapter, termed Feature Re-embedding (FR).

Specifically, to effectively capture discriminative temporal features, we introduce a critical component, Feature Re-embedding, within the temporal adapter. Feature Re-embedding is designed to enhance temporal representations by facilitating inter-frame collaboration to capture temporal dynamics, which ensures that relevant relations and implicit motion information are utilized for improved temporal modeling. As illustrated in the Fig.~\ref{fig:adapter} (b), given the reduced-dimensional spatial-temporal tokens $F\in\mathbb{R}^{T\times K\times C}$, we derive spatial tokens $\{F_s^i\in\mathbb{R}^{K\times C}\}_{i=1}^T$ for each temporal position and sequentially group these spatial embeddings based on a specified window size $w$. Consequently, the temporal sequence is partitioned into $k=\frac{T}{w}$ windows with fixed temporal resolution, resulting in grouped feature embeddings $\{G^j\in\mathbb{R}^{w\times K\times C}\}_{j=1}^k$. Subsequently, we establish long-term associations between independent frames within $G$ by performing a parameter-free reshaping operation, to reconfigure the temporal and channel dimensions. Specifically, for $G=\{F_s^1, F_s^2, ...,F_s^w\}$, we concatenate the spatial features at temporal positions $i$ and $i+\frac{w}{2}$ along the channel dimension, employing two GELU layers with a FC layer to generate the re-embedded features. The spatial pixel disparities across adjacent frames are interpreted as temporal information, facilitating the model to capture implicit motion dynamics and enhance temporal feature learning.

\begin{figure*}[t]
    \centering
    \begin{minipage}[t]{0.58\textwidth}
        \centering
        \begin{table}[H]
            \centering
            \caption{Overall comparison of SurgPETL based on three state-of-the-art methods, employing ViT-B and ViT-L pre-trained on five distinct large-scale datasets. \textbf{BOLD} indicates the best performance achieved by each method with varying pre-trained parameters.}
            \resizebox{\textwidth}{!}{
                \begin{tabular}{ c c | c c c c | c c c c }
                    	\toprule
    \multirow{2}{2.0cm}{\centering \textbf{Method}} & \multirow{2}{2cm}{\centering \textbf{Pre-trained \\ Parameters}} & \multicolumn{4}{c|}{\textbf{Unrelaxed Evaluation}} & \multicolumn{4}{c}{\textbf{Relaxed Evaluation}} \\
      \cmidrule(lr){3-6}  \cmidrule(lr){7-10}
     & & \textbf{Accuracy} & \textbf{Precision} & \textbf{Recall} & \textbf{Jaccard} & \textbf{Accuracy} & \textbf{Precision} & \textbf{Recall} & \textbf{Jaccard} \\
    \midrule
    \multirow{4}{2.0cm}{\centering \textbf{AIM \\ ViT-B} \\ (10.66M/ \\96.46M)}
    & ImagNet-21K  & 88.5$\pm$\scriptsize{6.9} & 83.1$\pm$\scriptsize{7.1} & 84.9$\pm$\scriptsize{6.9} & 71.9$\pm$\scriptsize{9.7} & 89.7$\pm$\scriptsize{7.0}  & 87.3$\pm$\scriptsize{4.4} & \textbf{89.0$\pm$\scriptsize{4.7}} & 76.8$\pm$\scriptsize{7.2}  \\
    & WIT-400M  & \textbf{89.8$\pm$\scriptsize{5.3}} & \textbf{85.3$\pm$\scriptsize{7.3}} & 83.0$\pm$\scriptsize{9.5} & 72.2$\pm$\scriptsize{12.0} & \textbf{91.0$\pm$\scriptsize{5.4}}  & \textbf{89.4$\pm$\scriptsize{5.5}} & 87.6$\pm$\scriptsize{6.0} & 77.5$\pm$\scriptsize{9.3}  \\
    & LAION-400M & 89.5$\pm$\scriptsize{6.1} & 85.0$\pm$\scriptsize{5.6} & \textbf{85.6$\pm$\scriptsize{7.4}} & \textbf{74.0$\pm$\scriptsize{9.7}} & 90.4$\pm$\scriptsize{6.2}  & 88.1$\pm$\scriptsize{4.2} & 88.5$\pm$\scriptsize{6.5} & 77.6$\pm$\scriptsize{8.1}  \\
    & LAION-2B & 89.3$\pm$\scriptsize{6.6} & \textbf{85.3$\pm$\scriptsize{5.0}} & 85.3$\pm$\scriptsize{7.0} & \textbf{74.0$\pm$\scriptsize{9.1}} & 90.3$\pm$\scriptsize{6.7}  & 88.9$\pm$\scriptsize{2.9} & 88.7$\pm$\scriptsize{5.1} & \textbf{78.2$\pm$\scriptsize{6.9}}  \\
    \midrule
   \multirow{4}{2.0cm}{\centering \textbf{AIM \\ ViT-L} \\ (37.86M/ \\ 341.16M)}
    & WIT-400M  & 91.3$\pm$\scriptsize{5.7} & 86.7$\pm$\scriptsize{6.4} & 88.6$\pm$\scriptsize{5.5} & 77.8$\pm$\scriptsize{8.6} & 92.0$\pm$\scriptsize{5.7}  & 89.7$\pm$\scriptsize{4.6} & 91.0$\pm$\scriptsize{4.7} & 81.0$\pm$\scriptsize{7.4}  \\
    & LAION-400M  & 90.2$\pm$\scriptsize{5.8} & 86.1$\pm$\scriptsize{6.7} & 86.4$\pm$\scriptsize{8.2} & 75.4$\pm$\scriptsize{10.0} & 91.1$\pm$\scriptsize{5.9}  & 88.9$\pm$\scriptsize{4.9} & 89.3$\pm$\scriptsize{7.2} & 78.9$\pm$\scriptsize{8.2}  \\
    & LAION-2B  & 91.2$\pm$\scriptsize{5.7} & 86.5$\pm$\scriptsize{8.0} & 87.7$\pm$\scriptsize{7.0} & 77.0$\pm$\scriptsize{10.0} & 92.1$\pm$\scriptsize{5.8}  & 89.7$\pm$\scriptsize{6.1} & 90.5$\pm$\scriptsize{5.6} & 80.5$\pm$\scriptsize{8.1} \\
    & SURGERY-23M  & \textbf{93.1$\pm$\scriptsize{5.4}} & \textbf{88.4$\pm$\scriptsize{6.1}} & \textbf{89.8$\pm$\scriptsize{6.5}} & \textbf{80.1$\pm$\scriptsize{9.1}} & \textbf{93.9$\pm$\scriptsize{5.4}}  & \textbf{91.2$\pm$\scriptsize{5.2}} & \textbf{91.9$\pm$\scriptsize{5.5}} & \textbf{83.4$\pm$\scriptsize{7.3}} \\
    \midrule
    \multirow{4}{2.0cm}{\centering \textbf{ST-Adapter \\ ViT-B} \\ (14.23M/ \\ 100.03M)}
    & ImagNet-21K  & \textbf{87.1$\pm$\scriptsize{7.1}} & \textbf{83.9$\pm$\scriptsize{4.3}} & \textbf{81.1$\pm$\scriptsize{9.5}} & \textbf{69.6$\pm$\scriptsize{9.8}} & \textbf{88.6$\pm$\scriptsize{7.3}}  & \textbf{88.4$\pm$\scriptsize{1.6}} & \textbf{86.0$\pm$\scriptsize{8.6} }& \textbf{75.0$\pm$\scriptsize{7.8}}  \\
    & WIT-400M  & 86.5$\pm$\scriptsize{7.3} & 82.4$\pm$\scriptsize{5.4} & 80.0$\pm$\scriptsize{9.2} & 67.6$\pm$\scriptsize{10.9} & 88.1$\pm$\scriptsize{7.7}  & 87.5$\pm$\scriptsize{3.5} & 85.7$\pm$\scriptsize{7.4} & 73.5$\pm$\scriptsize{9.1}  \\
    & LAION-400M & 85.8$\pm$\scriptsize{7.2} & 82.6$\pm$\scriptsize{5.1} & 76.9$\pm$\scriptsize{11.7} & 65.5$\pm$\scriptsize{11.8} & 87.5$\pm$\scriptsize{7.6}  & 88.0$\pm$\scriptsize{3.8} & 83.0$\pm$\scriptsize{8.8} & 71.8$\pm$\scriptsize{9.3}  \\
    & LAION-2B & 86.4$\pm$\scriptsize{6.5} & 82.2$\pm$\scriptsize{5.1} & 78.8$\pm$\scriptsize{11.3} & 66.5$\pm$\scriptsize{12.1} & 88.0$\pm$\scriptsize{6.8}  & 87.7$\pm$\scriptsize{3.8} & 84.8$\pm$\scriptsize{7.9} & 72.8$\pm$\scriptsize{9.3}  \\
    \midrule
   \multirow{4}{2.0cm}{\centering \textbf{ST-Adapter \\ ViT-L} \\ (37.90M / \\ 341.20M)}
    & WIT-400M  & 88.2$\pm$\scriptsize{7.3} & 84.9$\pm$\scriptsize{4.8} & 81.3$\pm$\scriptsize{10.0} & 70.5$\pm$\scriptsize{10.5} & 89.6$\pm$\scriptsize{7.5}  & 89.3$\pm$\scriptsize{3.0} & 86.0$\pm$\scriptsize{8.8} & 76.0$\pm$\scriptsize{8.7}  \\
    & LAION-400M  & 87.8$\pm$\scriptsize{7.0} & 84.1$\pm$\scriptsize{5.3} & 80.7$\pm$\scriptsize{9.9} & 69.6$\pm$\scriptsize{11.1} & 89.2$\pm$\scriptsize{7.3}  & 88.8$\pm$\scriptsize{2.6} & 85.7$\pm$\scriptsize{8.3} & 75.1$\pm$\scriptsize{9.1} \\
    & LAION-2B  & 86.9$\pm$\scriptsize{6.8} & 86.9$\pm$\scriptsize{6.8} & 78.6$\pm$\scriptsize{11.6} & 67.7$\pm$\scriptsize{11.5} & 88.4$\pm$\scriptsize{7.0}  & 88.4$\pm$\scriptsize{3.0} & 84.2$\pm$\scriptsize{9.7} & 73.5$\pm$\scriptsize{9.5} \\
    & SURGERY-23M  & \textbf{92.0$\pm$\scriptsize{5.6}} & \textbf{87.8$\pm$\scriptsize{6.1}} & \textbf{86.7$\pm$\scriptsize{8.0}} & \textbf{77.1$\pm$\scriptsize{10.2}} & \textbf{93.1$\pm$\scriptsize{5.5}}  & \textbf{91.4$\pm$\scriptsize{4.5}} & \textbf{90.0$\pm$\scriptsize{7.4}} & \textbf{81.5$\pm$\scriptsize{8.7}}  \\
    \midrule
    \multirow{4}{2.0cm}{\centering \textbf{Dual-Path \\ ViT-B} \\ (18.06M/ \\103.68M)}
    & ImagNet-21K  & 86.3$\pm$\scriptsize{7.8} & \textbf{81.0$\pm$\scriptsize{12.5}} &
    \textbf{73.7$\pm$\scriptsize{21.3}} & 
    \textbf{63.7$\pm$\scriptsize{22.6}} & 87.7$\pm$\scriptsize{7.7}  & \textbf{87.5$\pm$\scriptsize{8.6}} & 79.6$\pm$\scriptsize{17.1} & 69.0$\pm$\scriptsize{18.7} \\
    & WIT-400M  & 85.5$\pm$\scriptsize{8.3} & 79.3$\pm$\scriptsize{11.8} & 72.2$\pm$\scriptsize{19.6} & 62.8$\pm$\scriptsize{21.2} & 86.9$\pm$\scriptsize{8.1} & 87.1$\pm$\scriptsize{9.2} & 80.0$\pm$\scriptsize{16.5} & 69.2$\pm$\scriptsize{17.9} \\
    & LAION-400M  & 86.7$\pm$\scriptsize{7.0} & 79.0$\pm$\scriptsize{14.7} & 73.5$\pm$\scriptsize{21.5} & 62.9$\pm$\scriptsize{24.1} & 88.2$\pm$\scriptsize{6.8} & 85.9$\pm$\scriptsize{10.3} & 79.9$\pm$\scriptsize{16.8} & 68.7$\pm$\scriptsize{19.9} \\
    & LAION-2B  & \textbf{87.0$\pm$\scriptsize{6.9}} & 80.4$\pm$\scriptsize{13.4} & \textbf{73.7$\pm$\scriptsize{20.2}} & \textbf{63.7$\pm$\scriptsize{22.6}} & \textbf{88.5$\pm$\scriptsize{6.9}} & 87.2$\pm$\scriptsize{8.7} & \textbf{80.1$\pm$\scriptsize{15.2}} & \textbf{69.4$\pm$\scriptsize{18.4}} \\
    \midrule
   \multirow{4}{2.0cm}{\centering \textbf{Dual-Path \\ ViT-L} \\ (63.48M/ \\366.50M)}
    & WIT-400M  & 87.0$\pm$\scriptsize{6.9} & 80.4$\pm$\scriptsize{13.4} & 73.7$\pm$\scriptsize{20.2} & 63.7$\pm$\scriptsize{22.6} & 89.5$\pm$\scriptsize{6.9} & 87.5$\pm$\scriptsize{8.5} & 84.0$\pm$\scriptsize{10.2} & 72.8$\pm$\scriptsize{15.6} \\
    & LAION-400M  & 87.1$\pm$\scriptsize{7.1} & 81.1$\pm$\scriptsize{13.1} & 73.8$\pm$\scriptsize{20.5} & 63.9$\pm$\scriptsize{22.3} & 88.7$\pm$\scriptsize{6.9} & 87.3$\pm$\scriptsize{9.4} & 80.4$\pm$\scriptsize{15.5} & 69.9$\pm$\scriptsize{17.8} \\
    & LAION-2B  & 87.7$\pm$\scriptsize{6.0} & 81.4$\pm$\scriptsize{14.2} & 75.9$\pm$\scriptsize{22.1} & 65.3$\pm$\scriptsize{23.1} & 89.2$\pm$\scriptsize{5.8} & 87.0$\pm$\scriptsize{10.5} & 81.9$\pm$\scriptsize{17.6} & 71.1$\pm$\scriptsize{18.8} \\
    & SURGERY-23M  &  \textbf{90.6$\pm$\scriptsize{6.5}} & \textbf{83.6$\pm$\scriptsize{12.4}} & \textbf{80.7$\pm$\scriptsize{13.7}} & \textbf{71.4$\pm$\scriptsize{18.3}} & \textbf{92.1$\pm$\scriptsize{6.2}} & \textbf{90.4$\pm$\scriptsize{7.0}} & \textbf{87.1$\pm$\scriptsize{8.5}} & \textbf{77.5$\pm$\scriptsize{13.6}} \\
    \bottomrule
    \end{tabular}
            }
            \label{tab:benchmark}
        \end{table}
    \end{minipage}
    \hfill
    \begin{minipage}[t]{0.4\textwidth}
        \vspace{0pt}
        \includegraphics[width=\textwidth]{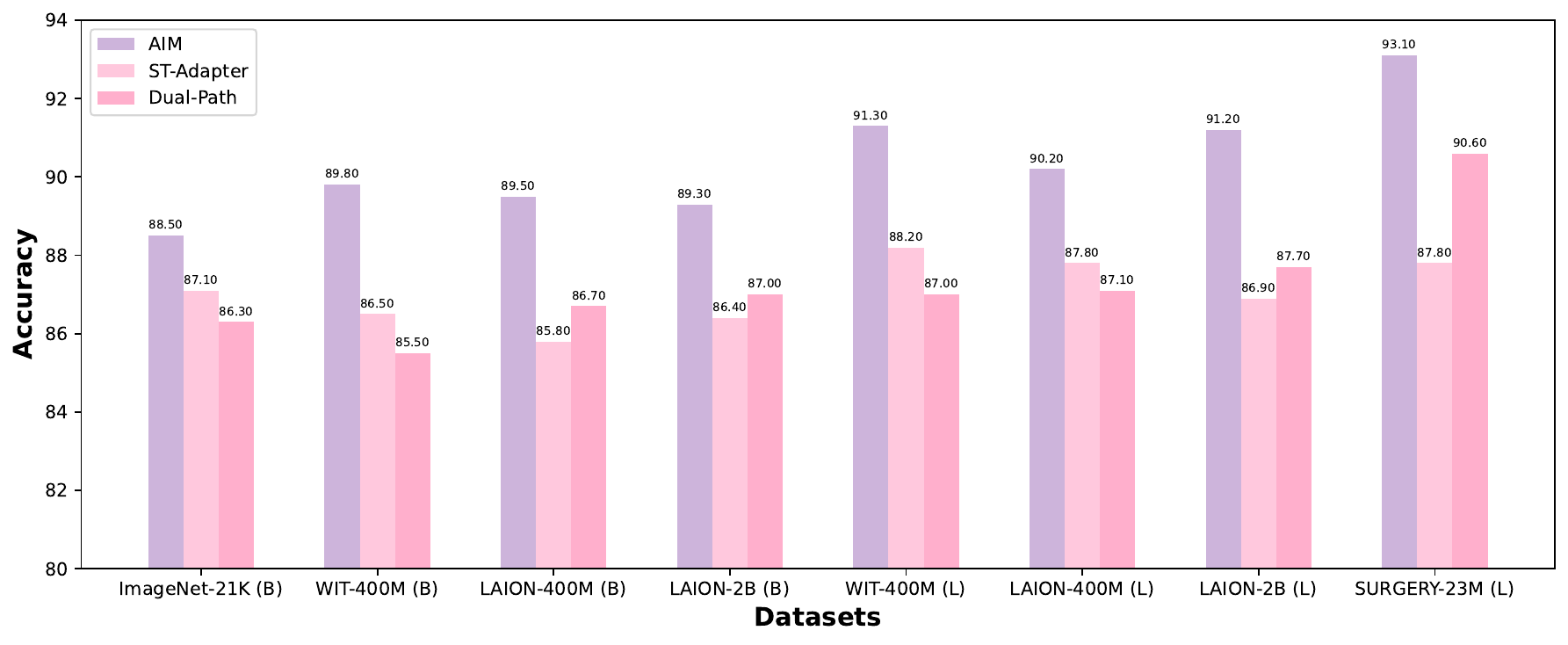}
        \caption{Performance of various methods with distinct pre-trained parameters and model scales (B: ViT-B; 
L: ViT-L).}
        \label{fig:5}
        \vspace{4mm}
        \includegraphics[width=\textwidth]{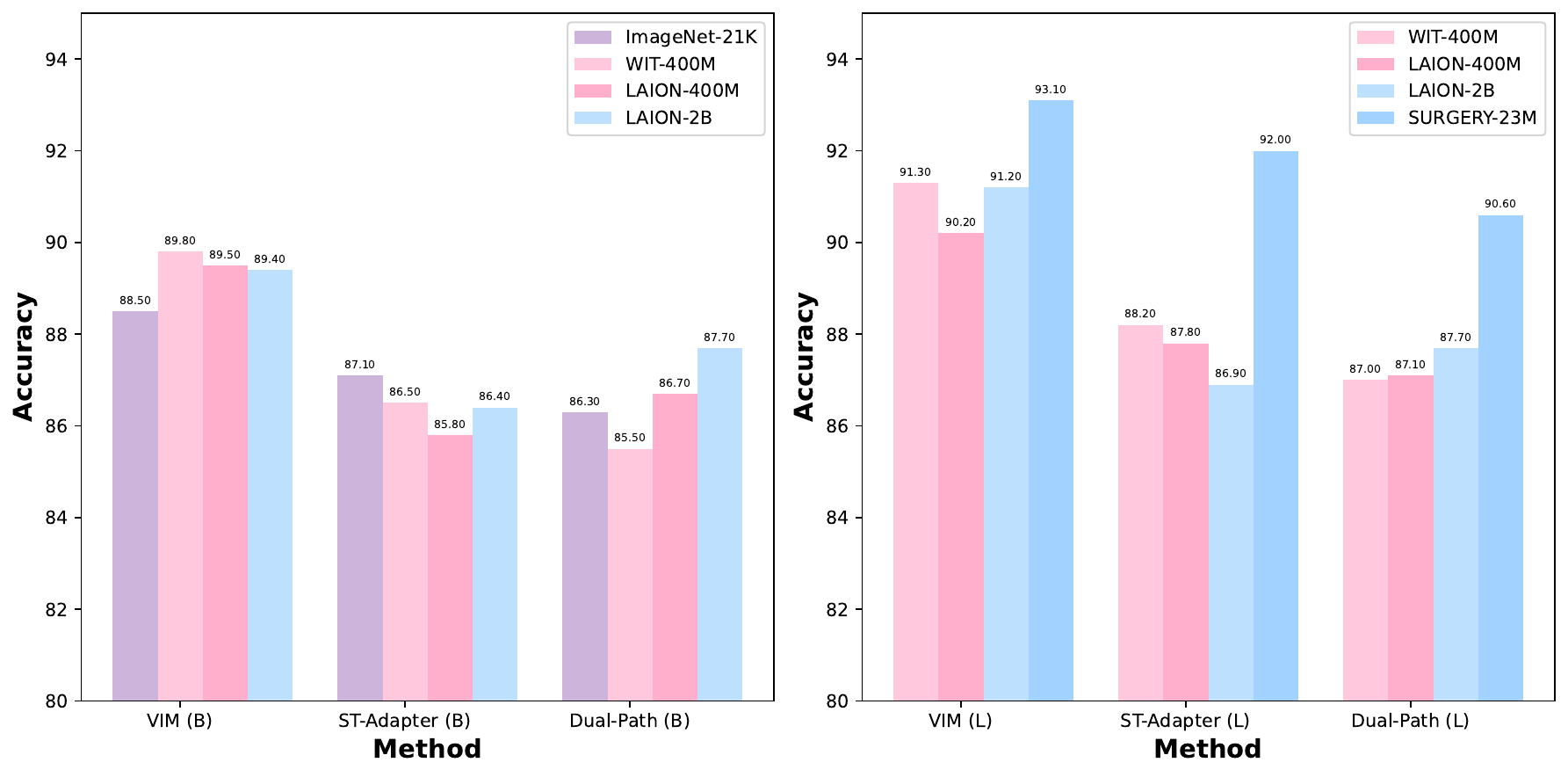}
        \caption{Comparison of various methods with distinct pre-trained parameters across two model scales.}
        \label{fig:6}
    \end{minipage}
\end{figure*}

\section{Experiments}

\subsection{Experimental Setup}
\subsubsection{Datasets}
To evaluate the performance of our proposed SurgPETL and STA, we conduct comparison experiments on three public challenging datasets: \textbf{\textit{Cholec80}}~\cite{endonet}, \textbf{\textit{Autolaparo}}~\cite{Autolaparo} and \textbf{\textit{Cataract-101}}~\cite{cataract101}. Cholec80 contains 80 cholecystectomy surgery videos, annotated with 7 distinct phase labels. These videos are officially divided into separate training and test sets, consisting of 40 videos each. Autolaparo contains 21 videos of laparoscopic hysterectomy with manual annotations of 7 surgical phases. Following~\cite{Autolaparo}, we split the dataset into 10 videos for training, 4 videos for validation and the remaining 7 videos for testing. Cataract-101 comprises of 101 cataract surgery video recordings, each annotated into 10 steps by surgeons. In accordance with previous study~\cite{GLSformer}, we perform five-fold cross validation with 50 videos for training, 10 videos for validation and 40 videos for testing. For fair comparisons, we sample all datasets into 1 fps and do not use any additional labels.
\subsubsection{Evaluation Metric}
Following the standard setting~\cite{TMRNet,LoViT}, we use four widely-used benchmark metrics: video-level accuracy, phase-level precision, phase-level recall and phase-level Jaccard. We employ both relaxed and unrelaxed evaluations for Cholec80, and only unrelaxed evaluation for Autolaparo and Cataract-101. In relaxed evaluation, predictions that fall within a 10-second window around the phase transition and correspond to neighboring phases are considered correct, even if they do not precisely match the ground truth.
\subsubsection{Implementation Details}
We initialize the shared weights using corresponding pre-trained weights, with the remaining layers being randomly initialized. We train all variants of the SurgPETL benchmark for 30 epochs with batch size as 64 on 4 NVIDIA RTX A6000 GPUs. We opt AdamW as optimizer, with $\beta_1$ as 0.9, $\beta_2$ as 0.999, weight decay as 5e-2 and initial learning rate as 3e-4. The learning rate is warmed up from 1e-6 in the first 3 epochs and then decays following a cosine schedule. The stochastic depth rate is 0.2 for both ViT-B and ViT-L. Following~\cite{AIM}, we utilize I3D~\cite{I3D} as the decoder for all variants of the SurgPETL benchmark, and the decoder employs learning rate 10 times of the backbone.  

In \textbf{Phase-I}, we evaluate the AIM and ST-Adapter using 8 frames ($T$ = 8), with a sampling interval of 4 ($R$ = 4), termed as $8\times4$. For Dual-Path, we utilize 8 frames with a sampling interval of 4 as the input of spatial adaption path, while 16 scaled frames with a sampling interval of 2 are transformed into a grid-like frameset for temporal adaptation path. Specifically, the 16 scaled frames are sequentially stacked in temporal order, forming a grid-like frameset that retains the dimensions of the original frame. In \textbf{phase-II}, unless stated otherwise, we employ a fixed configuration, 16 frames with a sampling interval of 4, for training and testing on both Autolaparo and Cataract-101 datasets. For the Cholec80 dataset, we utilize a training configuration with $T$ = 16 and $R$ = 4. In the ablation study, the same configuration is applied during testing, while for the evaluation, we adopt a testing configuration with $T$ = 16 and $R$ = 8 for higher performance. Note that we employ ViT-B backbone pre-trained on WIT-400M to maintain computational efficiency while ensuring promising performance.

\subsection{SurgPETL Benchmark}

\label{Sec:phase1}

\subsubsection{Comparison of three PEIVTL methods}
To facilitate fair and rigorous comparisons, we benchmark the SurgPETL with three advanced methods by fine-tuning on the Cholec80 dataset. As shown in Table~\ref{tab:benchmark} and Fig.~\ref{fig:5}, AIM exhibits superior performance compared to ST-Adapter and Dual-Path across various model scales and pre-trained parameters while utilizing fewer tuned parameters. For instance, using parameters pre-trained on ImageNet-21K, AIM consistently surpasses Dual-Path by 1.4\% and 1.1\% in accuracy of relaxed and unrelaxed evaluations, respectively. Notably, Dual-Path's performance is significantly inferior to that of AIM, which contradicts its efficacy in video action recognition~\cite{DUAL-PATH}. This discrepancy can be attributed to Dual-Path's proficiency in capturing global temporal features from a grid-like frameset for video-level tasks, which hinders its ability to effectively extract discriminative temporal features necessary for fine-grained frame-level tasks. Consequently, developing a temporal adapter that specifically targets the extraction of discriminative features is essential. Meanwhile, both ST-Adapter and Dual-Path achieve competitive results, highlighting the effectiveness of parameter-efficient image-to-surgical-video transfer learning paradigm.

\subsubsection{Effect of pre-trained parameters}
We examine the performance of all three methods on the Cholec80 dataset employing the ViT-B backbone pre-trained on four natural datasets: ImageNet-21K, WIT-400M, LAION-400M, and LAION-2B. Fig.~\ref{fig:6} illustrates that each model achieves optimal performance by leveraging different pre-trained parameters, revealing the effectiveness of natural pre-trained parameters. The performance difference between WIT-400M and LAION-400M demonstrates the sensitivity to varying data distributions. Additionally, increasing the volume of pre-trained data from 400M to 2B fails to yield improved performance for all three methods. Besides, when comparing variants based on diverse pre-trained parameters, the performance disparities among these variants remain minimal. Furthermore, we assess the performance of pre-trained parameters from distinct domains by utilizing the ViT-L backbone pre-trained on natural and surgical domains. Table~\ref{tab:benchmark} and Fig.~\ref{fig:6} indicate that all three methods achieve significant performance improvements across all metrics in both relaxed and unrelaxed evaluations when utilizing surgical pre-trained parameters compared to using natural pre-trained parameters. This implies that image features learned from surgical domain are intrinsically more expressive, enabling the model to seamlessly adapt surgical knowledge with minimal tuned parameters for fine-tuning. 

\begin{table}[t]
	\centering
	\caption{Comparison with the full fine-tuning counterpart TimeSformer on the Cholec80 dataset.}
    \resizebox{8.8cm}{!}{
	\begin{tabular}	{ c | c c c c | c c c c }
	\toprule
     \multirow{2}{1cm}{\centering \textbf{Setting}} &  \multicolumn{4}{c|}{\textbf{TimeSformer} (121.26M / 121.26M)} & \multicolumn{4}{c}{\textbf{SurgPETL-AIM} (10.66M / 96.46M)} \\
      \cmidrule(lr){2-5}  \cmidrule(lr){6-9}
     & Accuracy & Precision & Recall & Jaccard & Accuracy & Precision & Recall & Jaccard \\
    \midrule
    \rowcolor{mygray}
    \multicolumn{9}{c}{\textbf{Unrelaxed Evaluation}}\\
    \midrule
    8$\times$4 & 88.5$\pm$\scriptsize{5.7} & 82.9$\pm$\scriptsize{8.9} & \textbf{84.0$\pm$\scriptsize{9.7}} & 71.2$\pm$\scriptsize{12.2} & \textbf{89.8$\pm$\scriptsize{5.3}} & \textbf{85.3$\pm$\scriptsize{7.3}} & 83.0$\pm$\scriptsize{9.5} & \textbf{72.2$\pm$\scriptsize{12.0}}  \\
    12$\times$4 & 89.1$\pm$\scriptsize{7.0} & 84.1$\pm$\scriptsize{7.2} & 82.6$\pm$\scriptsize{14.6} & 71.4$\pm$\scriptsize{14.9} & \textbf{90.1$\pm$\scriptsize{6.5}} & \textbf{86.7$\pm$\scriptsize{6.4}} & \textbf{86.9$\pm$\scriptsize{6.6}} & \textbf{76.2$\pm$\scriptsize{8.7}}  \\
    16$\times$4 & 89.5$\pm$\scriptsize{5.8} & 85.4$\pm$\scriptsize{8.3} & 82.3$\pm$\scriptsize{13.9} & 71.5$\pm$\scriptsize{13.8} & \textbf{90.8$\pm$\scriptsize{6.4}} & \textbf{86.9$\pm$\scriptsize{6.8}} & \textbf{87.6$\pm$\scriptsize{6.1}} & \textbf{77.4$\pm$\scriptsize{9.4}}  \\
    \midrule
    \rowcolor{mygray}
    \multicolumn{9}{c}{\textbf{Relaxed Evaluation}}\\
    \midrule
    8$\times$4  & 89.7$\pm$\scriptsize{6.0} & 87.3$\pm$\scriptsize{7.0} & \textbf{88.4$\pm$\scriptsize{8.0}} & 76.1$\pm$\scriptsize{9.8} & \textbf{91.0$\pm$\scriptsize{5.4}}  & \textbf{89.4$\pm$\scriptsize{5.5}} & 87.6$\pm$\scriptsize{6.0} & \textbf{77.5$\pm$\scriptsize{9.3}} \\
    12$\times$4  & 90.5$\pm$\scriptsize{7.2}&
    88.7$\pm$\scriptsize{5.6} &
    87.4$\pm$\scriptsize{11.6} &
    76.7$\pm$\scriptsize{12.3} & \textbf{90.9$\pm$\scriptsize{6.5}} & \textbf{89.3$\pm$\scriptsize{4.6}} & \textbf{89.3$\pm$\scriptsize{6.3}} & \textbf{79.2$\pm$\scriptsize{7.0}}  \\
    16$\times$4  & 90.8$\pm$\scriptsize{6.1} &
    \textbf{89.7$\pm$\scriptsize{5.6}} &
    86.7$\pm$\scriptsize{10.4} &
    77.0$\pm$\scriptsize{10.4} & \textbf{91.6$\pm$\scriptsize{6.5}} & 89.6$\pm$\scriptsize{4.7} & \textbf{90.2$\pm$\scriptsize{4.9}} & \textbf{80.6$\pm$\scriptsize{7.4}}  \\
    \bottomrule
	\end{tabular}}
	\label{tab:comp}
\end{table}

\begin{table}[t]
	\centering
	\caption{Ablation analysis of SurgPETL-STA on the Autolaparo dataset. \textbf{Bold} indicates the highest performance in each training configuration.}
    \resizebox{8.8cm}{!}{
	\begin{tabular}	{ c c | c | c c c c }
	\toprule
     \multirow{2}{1cm}{\centering \textbf{Method}} & \multirow{2}{1.5cm}{\centering \textbf{Tuned/All Parameters}} & \multirow{2}{2.0cm}{\centering \textbf{Image-level \\ Accuracy}} &  \multicolumn{4}{c}{\textbf{Unrelaxed Evaluation}} \\
     \cmidrule(lr){4-7}
     &&& Accuracy & Precision & Recall & Jaccard \\
    \midrule
    \rowcolor{mygray}
    \multicolumn{7}{c}{\textbf{ViT-B / WIT-400M / 8 $\times$ 4}}\\
    \midrule
    \textbf{AIM} & 10.66M/96.46M & 81.12 & 81.27 & 78.44 & \textbf{73.10} & 61.36  \\
    \textbf{k=1} & 15.99M/101.79M & \textbf{82.25} & \textbf{82.60} & \textbf{78.91} & 70.35 & 60.50 \\
    \textbf{k=2} & 15.99M/101.79M & 82.13 & 82.40 & 77.47 & 70.84 & \textbf{61.60} \\
    \textbf{k=4} & 15.99M/107.12M & 81.82 & 82.21  & 76.58 & 69.41 & 59.59 \\
    \midrule
    \rowcolor{mygray}
    \multicolumn{7}{c}{\textbf{ViT-B / WIT-400M / 16 $\times$ 4}}\\
    \midrule
    \textbf{AIM} & 10.66M/96.46M & 81.54 & 81.62 & 76.19 & 73.96 & 61.81 \\
    \textbf{k=1} & 15.99M/101.79M & 84.08 & 84.39 & 79.51 & 73.65 & 64.73 \\
    \textbf{k=2} & 15.99M/101.79M & \textbf{84.79} & \textbf{85.02} & \textbf{82.55} & \textbf{74.69} & \textbf{66.06} \\
    \textbf{k=4} & 15.99M/107.12M & 83.93 & 84.15 & 77.68 & 74.11 & 64.70 \\
    \textbf{k=8} & 15.99M/107.12M & 83.87 & 84.07 & 77.95 & 74.23 & 64.77 \\
    \bottomrule
	\end{tabular}}
	\label{ab:autolaparo}
\end{table}

\begin{table}[t]
	\centering
	\caption{Ablation analysis of SurgPETL-STA on the Cholec80 dataset. \textcolor{red}{Red} and \textcolor{blue}{Blue} indicates the performance improvements and degradations relative to AIM in each training configuration.}
    \resizebox{8.8cm}{!}{
	\begin{tabular}	{ c c | c | c c c | c c c }
	\toprule
     \multirow{2}{1cm}{\centering \textbf{Method}} & \multirow{2}{1.0cm}{\centering \textbf{Tuned/All \\ Parameters}} & \multirow{2}{2.0cm}{\centering \textbf{Image-level \\ Accuracy}} &  \multicolumn{3}{c|}{\textbf{Unrelaxed Evaluation}} & \multicolumn{3}{c}{\textbf{Relaxed Evaluation}} \\
      \cmidrule(lr){4-6}  \cmidrule(lr){7-9}
     &&& Accuracy & F1 & Jaccard & Accuracy & F1 & Jaccard \\
    \midrule
    \rowcolor{mygray}
    \multicolumn{9}{c}{\textbf{ViT-B / WIT-400M / 8 $\times$ 4}}\\
    \midrule
    \textbf{AIM} & 10.66M/96.46M & 89.6 & 89.8 & 82.3 & 72.2 & 91.0  & 87.7 & 77.5  \\
    \textbf{k=4} & 15.99M/101.79M & 89.9 & 90.2  & 84.4 & 75.2 & 91.1 & 88.0 & 78.7  \\
    \rowcolor{dino}
    $\Delta$ & - & \textbf{\textcolor{red}{+0.3}} & \textbf{\textcolor{red}{+0.4}}  & \textbf{\textcolor{red}{+2.1}} & \textbf{\textcolor{red}{+3.0}} & \textbf{\textcolor{red}{+0.1}} & \textbf{\textcolor{red}{+0.3}} & \textbf{\textcolor{red}{+1.2}}  \\
     \textbf{k=2} & 15.99M/101.79M & 89.9 & 90.3  & 84.3 & 74.6 & 91.4 & 88.6 & 79.2  \\
    \rowcolor{dino}
    $\Delta$ & - & \textbf{\textcolor{red}{+0.3}} & \textbf{\textcolor{red}{+0.5}}  &  \textbf{\textcolor{red}{+2.0}} & \textbf{\textcolor{red}{+2.4}} & \textbf{\textcolor{red}{+0.4}} & \textbf{\textcolor{red}{+0.9}} & \textbf{\textcolor{red}{+1.7}}  \\
     \textbf{k=2/4} & 21.32M/107.12M & 89.6 & 90.0  & 84.0 & 74.5 & 90.9 & 87.9 & 78.4  \\
    \rowcolor{dino}
    $\Delta$ & - & \textbf{\textcolor{red}{+0.0}} & \textbf{\textcolor{red}{+0.2}}  &  \textbf{\textcolor{red}{+1.7}} & \textbf{\textcolor{red}{+2.3}} &  \textbf{\textcolor{blue}{-0.1}} & \textbf{\textcolor{red}{+0.2}} & \textbf{\textcolor{red}{+0.9}}  \\
    \midrule
    \rowcolor{mygray}
    \multicolumn{9}{c}{\textbf{ViT-B / WIT-400M / 16 $\times$ 4}}\\
    \midrule
    \textbf{AIM} & 10.66M/96.46M & 90.2 & 90.8 & 85.7 & 77.4 & 91.6  & 89.0 & 80.6  \\
    \textbf{k=8} & 15.99M/101.79M & 90.7 & 91.2 & 85.7 & 77.5 & 92.2  & 89.7 & 81.6  \\
    \rowcolor{dino}
    $\Delta$ & -  & \textbf{\textcolor{red}{+0.5}} & \textbf{\textcolor{red}{+0.4}} & \textbf{\textcolor{red}{+0.0}} & \textbf{\textcolor{red}{+0.1}} &  \textbf{\textcolor{red}{+0.6}} & \textbf{\textcolor{red}{+0.7}} & \textbf{\textcolor{red}{+1.0}} \\
    \textbf{k=2} & 15.99M/101.79M & 90.7 & 91.2 & 86.0 & 77.8 & 92.2 & 89.9 & 81.7  \\
    \rowcolor{dino}
    $\Delta$ & -  & \textbf{\textcolor{red}{+0.5}} & \textbf{\textcolor{red}{+0.4}} & \textbf{\textcolor{red}{+0.3}} & \textbf{\textcolor{red}{+0.4}} & \textbf{\textcolor{red}{+0.6}} & \textbf{\textcolor{red}{+0.9}} & \textbf{\textcolor{red}{+1.1}} \\
    \textbf{k=2/8} & 21.32M/107.12M & 90.9 & 91.3 & 86.3 & 78.1 & 92.0 & 89.4 & 81.0  \\
    \rowcolor{dino}
    $\Delta$ & -  & \textbf{\textcolor{red}{+0.7}} & \textbf{\textcolor{red}{+0.5}} & \textbf{\textcolor{red}{+0.6}} & \textbf{\textcolor{red}{+0.7}} & \textbf{\textcolor{red}{+0.4}} & \textbf{\textcolor{red}{+0.4}} & \textbf{\textcolor{red}{+0.4}} \\
    \bottomrule
	\end{tabular}}
	\label{ab:cholec80}
\end{table}

\subsubsection{Effect of model scales}
Fig.~\ref{fig:6} shows that we implement several variants of each method using ViT-B and ViT-L as backbones. The ViT-L generates better spatial features for temporal modeling, leading it to consistently outperform the ViT-B variant across all metrics based on different parameters.

\subsubsection{Fully fine-tuning vs PEIVTL} Furthermore, we assess the disparity between a fully fine-tuned counterpart TimeSformer~\cite{TimeSFormer} and AIM, thereby evaluating the quality of representations acquired through fully fine-tuning and transfer learning. For fair comparisons, we employ the ViT-B pre-trained on WIT-400M as backbone with three distinct training settings. Table~\ref{tab:comp} demonstrates that AIM surpasses the TimeSformer across nearly all metrics in both relaxed and unrelaxed evaluations with only 10.66M tuned parameters. Our proposed SurgPETL can avoid the risks of degrading or collapsing the foundation model that full-parameter fine-tuning might entail.
 
\subsection{SurgPETL-STA}
To examine the effectiveness of our proposed SurgPETL-STA,
we conduct comprehensive experiments including ablation study and thorough comparisons on three benchmark public challenging datasets spanning distinct surgical procedures. For fair and effective comparisons with existing methods, we implement SurgPETL-STA using the ViT-B backbone pre-trained on the WIT-400M dataset as shown in Fig.~\ref{fig:pipeline}.
\subsubsection{Ablation Study}
\label{ab}
To investigate the impact of the proposed STA module on performance, we employ various values of $k$ to implement distinct formations of temporal modeling, categorized as follows: (1) For $k=1$, STA acquires comprehensive temporal information across entire resolution, facilitating long-term information extraction to enhance global temporal comprehension. (2) For $1 \textless k \textless \frac{T}{2}$, the focus transitions to medium-term temporal patterns, capturing intricate inter-frame dynamics and achieving a balance between local and global information. (3) For $k=\frac{T}{2}$, STA consolidates short-term contexts, highlighting detailed motion nuances and improving local temporal consistency. Specifically, we employ AIM as a baseline and subsequently introduce each variants of STA into the baseline to assess its efficacy. As demonstrated in Table~\ref{ab:autolaparo}, extensive experiments on the AutoLaparo dataset reveal that all variants consistently outperform AIM across all metrics in two distinct training settings. SurgPETL-STA demonstrates superior performance by effectively learning temporal information within a sequence of predefined resolution 8, specifically utilizing $k=1$ for $T$ = 8 and $k=2$ for $T$ = 16. 

Furthermore, we conduct ablation study on the Cholec80 dataset using two distinct training configurations. We evaluate standard variants with $k=2$ and $k=4$ for the $8\times4$ training configuration and variants with $k=2$ and $k=8$ for the $16\times4$ training configuration. Additionally, we also implement variants that employ two separate temporal adapters, each with distinct values of $k$, to capture both short-term and global-term temporal information. As illustrated in Table~\ref{ab:cholec80}, all variants outperform AIM in nearly all metrics in both relaxed and unrelaxed evaluations. Notably, while the variants with $k=2/4$ or $k=2/8$ incorporate additional tuned parameters, they fail to consistently yield substantial performance improvements across all training configurations. In conclusion, AIM's performance is inevitably compromised due to deviations from its original spatial modeling capabilities, whereas the proposed STA module effectively  mitigates this limitation.

\begin{table}[t]
	\centering
	\caption{Overall comparisons with the state-of-the-art methods on the Cholec80 dataset. FF means full fine-tuning paradigm and PETL indicates end-to-end parameter-efficient image-to-surgical-video transfer learning.}
 \resizebox{8.8cm}{!}{
	\begin{tabular}	{ c c c c c c}
	\toprule
    \multirow{2}{*}{\textbf{Method}} & \multirow{2}{*}{\textbf{Paradigm}} & \textbf{Video-level Metric} & \multicolumn{3}{c}{\textbf{Phase-level Metric}} \\
    \cmidrule(lr){3-3} \cmidrule(lr){4-6}
    & & \textbf{Accuracy $\uparrow$} & \textbf{Precision $\uparrow$} & \textbf{Recall $\uparrow$} & \textbf{Jaccard $\uparrow$} \\
    \midrule
    \rowcolor{mygray}
    \multicolumn{6}{c}{\textbf{Relaxed Evaluation}}\\
    \midrule
    EndoNet~\cite{endonet} & FF &
    81.7 $\pm$ 4.2 & 
    73.7 $\pm$ 16.1 &
    79.6 $\pm$ 7.9 &
    - \\
    MTRCNet-CL~\cite{MTRCNet-CL} & FF & 
    89.2 $\pm$ 7.6 & 
    86.9 $\pm$ 4.3 & 
    88.0 $\pm$ 6.9 & 
    - \\
    SV-RCNet~\cite{SV-RCNet} &  FF & 
    85.3 $\pm$ 7.3 & 
    80.7 $\pm$ 7.0 & 
    83.5 $\pm$ 7.5 & 
    - \\
    OHFM~\cite{OHFM} & FF & 
    87.3 $\pm$ 5.7 & 
    - & 
    - & 
    67.0 $\pm$ 13.3 \\
    TeCNO~\cite{TeCNO} & FF & 
    88.6 $\pm$ 7.8 & 
    86.5 $\pm$ 7.0 & 
    87.6 $\pm$ 6.7 & 
    75.1 $\pm$ 6.9 \\
    TMRNet~\cite{TMRNet} & FF & 
    90.1 $\pm$ 7.6 & 
    90.3 $\pm$ 3.3 & 
    89.5 $\pm$ 5.0 & 
    79.1 $\pm$ 5.7 \\
    Trans-SVNet~\cite{Trans-SVNet} & FF & 
    90.3 $\pm$ 7.1 & 
    90.7 $\pm$ 5.0 & 
    88.8 $\pm$ 7.4 & 
    79.3 $\pm$ 6.6 \\
    LoViT~\cite{LoViT} & FF & 
    92.4 $\pm$ 6.3 & 
    89.9 $\pm$ 6.1 & 
    90.6 $\pm$ 4.4 & 
    81.2 $\pm$ 9.1 \\
    SKiT~\cite{SKiT} & FF & 
    \textbf{93.4 $\pm$ 5.2} & 
    90.9 & 
    91.8 & 
    82.6 \\
    \midrule
    \rowcolor{dino}
    SurgPETL-STA & PETL & 
    \textbf{93.4 $\pm$ 6.0} & 
    \textbf{91.4 $\pm$ 5.0} & 
    \textbf{92.7 $\pm$ 5.7} & 
    \textbf{83.8 $\pm$ 9.0} \\
    \midrule
    \rowcolor{mygray}
    \multicolumn{6}{c}{\textbf{Unrelaxed Evaluation}}\\
    \midrule
    Trans-SVNet~\cite{Trans-SVNet} & FF & 
    89.1 $\pm$ 7.0 & 
    84.7 & 
    83.6 & 
    72.5 \\
    AVT~\cite{AVT} & FF & 
    86.7 $\pm$ 7.6 & 
    77.3 & 
    82.1 & 
    66.4 \\
    LoViT~\cite{LoViT} & FF & 
    91.5 $\pm$ 6.1 & 
    83.1 & 
    86.5 & 
    74.2  \\
    SKiT~\cite{SKiT} & FF & 
    \textbf{92.5 $\pm$ 5.1} & 
    84.6 & 
    88.5 & 
    76.7 \\
    \midrule
    \rowcolor{dino}
    SurgPETL-STA & PETL & 
    92.3 $\pm$ 5.9 & 
    \textbf{87.2 $\pm$ 8.3} & 
    \textbf{88.9 $\pm$ 8.9} & 
    \textbf{79.1 $\pm$ 12.5} \\
    \bottomrule
	\end{tabular}}
	\label{tab:cholec80}
\end{table}

\begin{table}[t]
	\centering
	\caption{Overall comparisons with the state-of-the-art methods on the Autolaparo dataset.}
 \resizebox{8.8cm}{!}{
	\begin{tabular}	{ c c c c c c}
	\toprule
    \multirow{2}{*}{\textbf{Method}} & \multirow{2}{*}{\textbf{Paradigm}} & \textbf{Video-level Metric} & \multicolumn{3}{c}{\textbf{Phase-level Metric}} \\
    \cmidrule(lr){3-3} \cmidrule(lr){4-6}
    & & \textbf{Accuracy $\uparrow$} & \textbf{Precision $\uparrow$} & \textbf{Recall $\uparrow$} & \textbf{Jaccard $\uparrow$} \\
    \midrule
    SV-RCNet~\cite{SV-RCNet} &  FF & 
    75.6 & 
    64.0 & 
    59.7 & 
    47.2 \\
    TMRNet~\cite{TMRNet} & FF & 
    78.2 & 
    66.0 & 
    61.5 & 
    49.6 \\
    TeCNO~\cite{TeCNO} & FF & 
    77.3 & 
    66.9 & 
    64.6 & 
    50.7 \\
    Trans-SVNet~\cite{Trans-SVNet} & FF & 
    78.3 & 
    64.2 & 
    62.1 & 
    50.7 \\
    AVT~\cite{AVT} & FF & 
    77.8 & 
    68.0 & 
    62.2 & 
    50.7 \\
    LoViT~\cite{LoViT} & FF & 
    81.4 $\pm$ 7.6 & 
    \textbf{85.1} & 
    65.9 & 
    56.0 \\
    SKiT~\cite{SKiT} & FF & 
    82.9 $\pm$ 6.8 & 
    81.8 & 
    70.1 & 
    59.9 \\
    \midrule
    \rowcolor{dino}
    SurgPETL-STA & PETL & 
    \textbf{85.0 $\pm$ 6.3} & 
    82.6 & 
    \textbf{74.7} & 
    \textbf{66.1} \\
    \bottomrule
	\end{tabular}}
	\label{tab:autolaparo}
\end{table}

\subsubsection{Evaluation on the Cholec80 dataset} We further evaluate the performance of SurgPETL-STA against state-of-the-art methods on the Cholec80 dataset. The results of all compared methods are presented in Table~\ref{tab:cholec80}. Our proposed SurgPETL-STA is comparable with recent two-stage full fine-tuning methods across all metrics in both relaxed and unrelaxed evaluations. Our proposed SurgPETL-STA significantly outperforms the best performance SkiT~\cite{SKiT} across all phase-level metrics, with particularly significant improvements in the unrelaxed evaluation.

\subsubsection{Evaluation on the Autolaparo dataset}  
As illustrated in Table~\ref{tab:autolaparo}, we compare the performance of our proposed method with full fine-tuning methods on the more challenging Autolaparo dataset. SurgPETL-STA significantly outperforms all two-stage full fine-tuning methods and demonstrates substantial performance improvements across most metrics, while fine-tuning only a fraction of the parameters. Specifically, SurgPETL-STA surpasses the best performance SkiT~\cite{SKiT} under all metrics, achieving gains of 4.6\% and 6.2\% in terms of Recall and Jaccard. For more challenging datasets that rely on effective temporal modeling, such as AutoLaparo, SurgPETL-STA effectively and efficiently captures robust temporal information, resulting in significantly improved performance.

\subsubsection{Evaluation on the Cataract-101 dataset}   
Table~\ref{tab:cat} illustrates the results of all compared methods on the Cataract-101 dataset. Our proposed SurgPETL-STA demonstrates performance comparable to the state-of-the-art method~\cite{GLSformer} across all metrics. For small-scale datasets with limited annotated samples, SurgPETL-STA effectively tackles the overfitting challenge and achieves significantly improved performance.

\begin{table}[t]
	\centering
	\caption{Overall comparisons with the state-of-the-art methods on the Cataract-101 dataset.}
 \resizebox{8.8cm}{!}{
	\begin{tabular}	{ c c c c c c}
	\toprule
    \multirow{2}{*}{\textbf{Method}} & \multirow{2}{*}{\textbf{Paradigm}} & \textbf{Video-level Metric} & \multicolumn{3}{c}{\textbf{Phase-level Metric}} \\
    \cmidrule(lr){3-3} \cmidrule(lr){4-6}
    & & \textbf{Accuracy $\uparrow$} & \textbf{Precision $\uparrow$} & \textbf{Recall $\uparrow$} & \textbf{Jaccard $\uparrow$} \\
    \midrule
    SV-RCNet~\cite{SV-RCNet} &  FF & 
    86.13 $\pm$ 0.91 & 
    84.96 $\pm$ 0.94& 
    76.61 $\pm$ 1.18& 
    66.51 $\pm$ 1.30\\
    OHFM~\cite{TMRNet} & FF & 
    87.82 $\pm$ 0.71 & 
    85.37 $\pm$ 0.78 & 
    78.29 $\pm$ 0.81 & 
    69.01 $\pm$ 0.93\\
    TeCNO~\cite{TeCNO} & FF & 
    88.26 $\pm$ 0.92 & 
    86.03 $\pm$ 0.83 & 
    79.52 $\pm$ 0.90 & 
    70.18 $\pm$ 1.15 \\
    TMRNet~\cite{TMRNet} & FF &
    89.68 $\pm$ 0.76 & 
    85.09 $\pm$ 0.72 & 
    82.44 $\pm$ 0.75 & 
    71.83 $\pm$ 0.91\\
    Trans-SVNet~\cite{Trans-SVNet} & FF & 
    89.45 $\pm$ 0.88 & 
    78.51 $\pm$ 1.42& 
    75.62 $\pm$ 1.83& 
    64.77 $\pm$ 1.97\\
    GLSFormer~\cite{SKiT} & FF & 
    92.91 $\pm$ 0.67 & 
    90.04 $\pm$ 0.71 & 
    89.45 $\pm$ 0.79 & 
    81.89 $\pm$ 0.92 \\
    \midrule
    \rowcolor{dino}
    SurgPETL-STA & PETL & 
    \textbf{94.5 $\pm$ 0.11} & 
    \textbf{92.8 $\pm$ 0.10} & 
    \textbf{93.44 $\pm$ 0.16} & 
    \textbf{86.5 $\pm$ 0.07} \\
    \bottomrule
	\end{tabular}}
	\label{tab:cat}
\end{table}
 
\section{Conclusions}

To tackle the challenges posed by the high computational load, the limited availability of surgical video data, and the necessity for comprehensive spatial-temporal modeling, we delve into a novel problem of efficiently adapting pre-trained image models to specialize in fine-grained surgical phase recognition, termed as parameter-efficient image-to-surgical-video transfer learning. By conducting two-phase investigation, we establish a benchmark called SurgPETL through extensive experiments, and propose the Spatial-Temporal Adaption (STA) module for further performance improvements. Extensive experiments on three public challenging datasets verify that our proposed SurgPETL-STA outperforms existing parameter-efficient alternatives and state-of-the-art surgical phase recognition methods. SurgPETL-STA allows a pre-trained image model without temporal knowledge to analyze dynamic video content with minimal parameter overhead, which has the potential to leverage more robust image foundation models and generalize well to other surgical downstream tasks.

\bibliographystyle{IEEEtran}
\bibliography{SurgPETL}

\end{document}